\documentclass[10pt,twocolumn,letterpaper]{article}
\title{Learning Deep Representations of Fine-Grained Visual Descriptions}

\usepackage{cvpr}
\usepackage{times}
\usepackage{epsfig}
\usepackage{graphicx}
\usepackage{amsmath}
\usepackage{amssymb}
\usepackage{multirow}
\usepackage{caption}
\usepackage{subcaption}
\usepackage{verbatim}
\usepackage[pagebackref=true,breaklinks=true,letterpaper=true,colorlinks,bookmarks=false]{hyperref}

 \cvprfinalcopy 

\def\cvprPaperID{802} 

\ifcvprfinal\fi
\begin{document}
\author{Scott Reed$^1$, Zeynep Akata$^2$, Honglak Lee$^1$ and Bernt Schiele$^2$ \vspace{4mm} \\ 
{
\begin{tabular}{cp{0.5cm}c}
$^1$University of Michigan & & $^2$Max-Planck Institute for Informatics
\end{tabular}
}
}
\maketitle

\begin{abstract}
State-of-the-art methods for zero-shot visual recognition formulate learning as a joint embedding problem of images and side information. In these formulations the current best complement to visual features are attributes:  manually-encoded vectors describing shared characteristics among categories. 
Despite good performance, attributes have limitations: (1) finer-grained recognition requires commensurately more attributes, and (2) attributes do not provide a natural language interface.
We propose to overcome these limitations by training neural language models from scratch; \ie without pre-training and only consuming words and characters. Our proposed models train end-to-end to align with the fine-grained and category-specific content of images. Natural language provides a flexible and compact way of encoding only the salient visual aspects for distinguishing categories. By training on raw text, our model can do inference on raw text as well, providing humans a familiar mode both for annotation and retrieval. Our model achieves strong performance on zero-shot text-based image retrieval and significantly outperforms the attribute-based state-of-the-art for zero-shot classification on the Caltech-UCSD Birds 200-2011 dataset.
\end{abstract}
%
%
\section{Introduction}
\label{sec:intro}
A key challenge in image understanding is to correctly relate natural language concepts to the visual content of images.
In recent years there has been significant progress in learning visual-semantic embeddings, \eg for zero-shot learning~\cite{PPH09,RSS11,LNH13,NMBSSFCD13,FCSBM13,SGSBMN13,ARWLS15} and automatically generating image captions for general web images~\cite{KPD11,OKB11,VTBE15,KL15,DHGRVSD15}. These methods have harnessed large image and text datasets~\cite{ILSVRC15,young2014image,lin2014microsoft}, as well as advances in deep neural networks for image and language modeling, already enabling powerful new applications such as auto-captioning images for blind users on the web~\cite{fb2015caption}.

\begin{figure}[t]
\vspace{0.3in}
\begin{center}
\includegraphics[width=0.98\columnwidth]{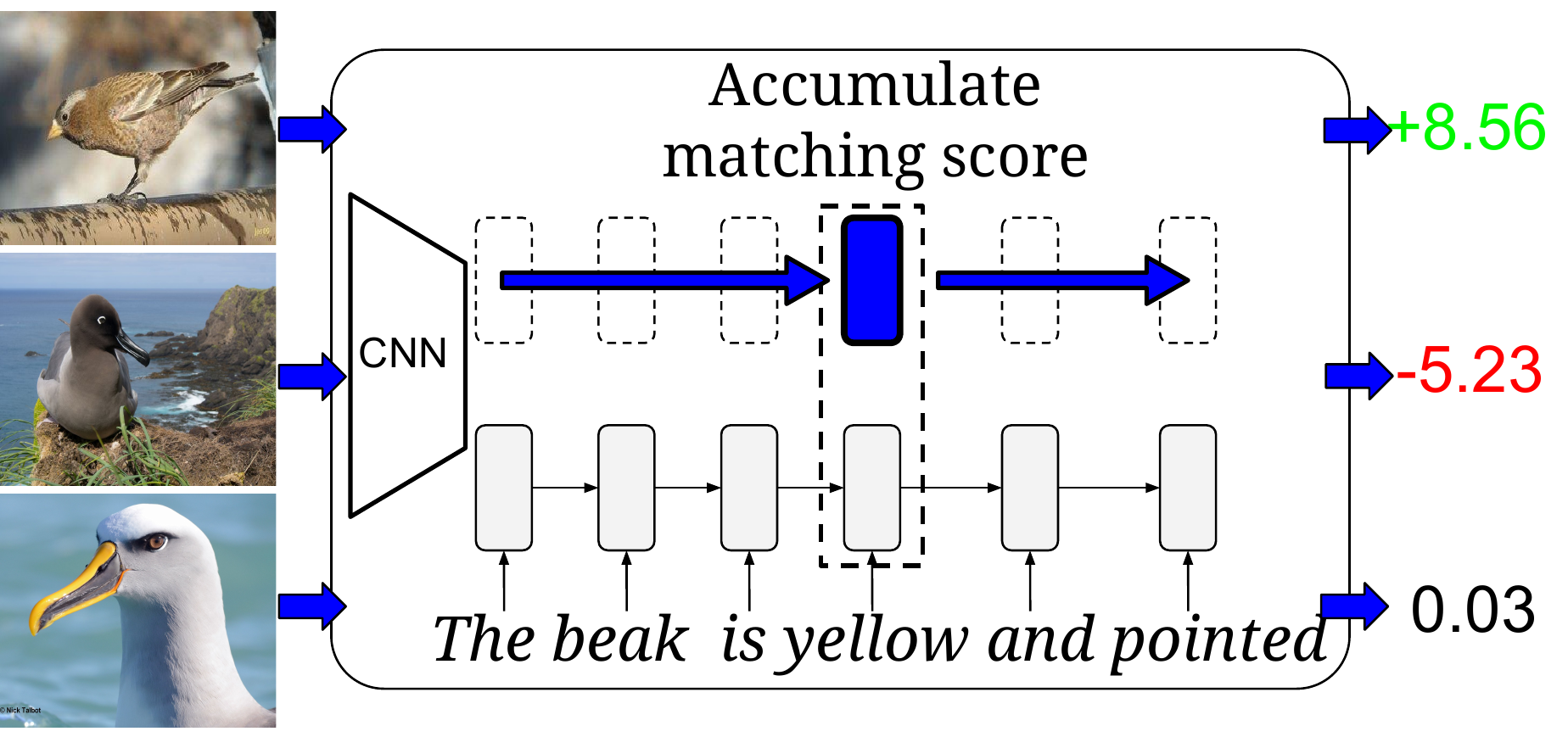}
\vspace{-0.1in}
\caption{Our model learns a scoring function between images and text descriptions. A word-based LSTM is shown here, but we also evaluate several alternative models.}
\label{fig:tcnn_model}
\end{center}
\vspace{-0.3in}
\end{figure}

Despite these advances, the problem of relating images and text is still far from solved.
In particular for the fine-grained regime~\cite{CaltechUCSDBirdsDataset,DPCG12,DKF13,ZDGD14}, where images of different classes have only subtle distinctions, sophisticated language models have not been employed, perhaps due to the scarcity of large and high-quality training data.
For instance on the Caltech-UCSD birds database (CUB)~\cite{CaltechUCSDBirdsDataset}, previous zero-shot learning approaches~\cite{FHXFG14,ARWLS15,ba2015predicting} have used human-encoded attributes~\cite{LNH13}, or simplified language models such as bag-of-words~\cite{H54}, WordNet-hierarchy-derived features~\cite{WordNet}, and neural word embeddings such as Word2Vec~\cite{MSCCD13} and GloVE~\cite{PSM14}.

Previous text corpora used for fine-grained label embedding were either very large but not visually focused, \eg the entire \texttt{wikipedia}, or somewhat visually relevant but very short, \eg the subset of \texttt{wikipedia} articles  that are related to birds.
Furthermore, these wikis do not provide enough aligned images and text to train a high-capacity sentence encoder.
Given the data limitations, previous text embedding methods work surprisingly well for zero-shot visual recognition, but there remains a large gap between the text embedding methods and human-annotated attributes ($28.4\%$ vs $50.1\%$ average top-1 per-class accuracy on CUB~\cite{ARWLS15}).

In order to close the performance gap between text embeddings and human-annotated attributes for fine-grained visual recognition, we hypothesize that higher-capacity text models are required.
However, more sophisticated text models would in turn require more training data, in particular aligned images and multiple visual descriptions per image for each fine-grained category.
These descriptions would support both zero-shot image recognition and zero-shot image retrieval, which are strong measures of the generalization ability of both image and text models.

Our contributions in this work are as follows.
First, we collected two datasets of fine-grained visual descriptions: one for the Caltech-UCSD birds dataset, and another for the Oxford-102 flowers dataset~\cite{OxfordFlowersDataset}.
Both our data and code will be made available.
Second, we propose a novel extension of structured joint embedding~\cite{ARWLS15}, and show that it can be used for end-to-end training of deep neural language models.
It also dramatically improves zero-shot retrieval performance for all models.
Third, we evaluate several variants of word- and character-based neural language models, including our novel hybrids of convolutional and recurrent networks for text modeling.
We demonstrate significant improvements over the state-of-the-art on CUB and Flowers datasets in both zero-shot recognition and retrieval.
%

%
%
%

\section{Related work}
\label{sec:references}
Over the past several years, advances in deep convolutional networks~\cite{KSH12,decaf,szegedy2015going} have driven rapid progress in general-purpose visual recognition on large-scale benchmarks such as ImageNet~\cite{DDS09}.
The learned features of these networks have proven transferable to many other problems~\cite{oquab2014learning}.
However, a remaining challenge is fine-grained image classification~\cite{CaltechUCSDBirdsDataset,DPCG12,DKF13,ZDGD14}, \ie classifying objects of many visually similar classes.
The difficulty is increased by the lack of extensive labeled images~\cite{PPH09,RSS11,LNH13,NMBSSFCD13,FCSBM13,SGSBMN13}, which for fine-grained data sets may even require annotation by human experts.

The setting we study in this work is both fine-grained and \emph{zero-shot}, \eg we want to do fine-grained classification of previously unseen categories of birds and flowers.
This problem is not as contrived as it may at first seem:
good performance would strongly indicate the generalization ability of image and text features; in particular that our visual description embeddings represent well the fine-grained visual \emph{concepts} in images, rather than over-fitting to known categories.
Strong performance metrics for visual-semantic models are especially apropos because of the risk of overfitting recent high-capacity captioning models, \eg memorizing (and possibly regurgitating) training captions.

We compare to previous work on zero-shot recognition, and also report zero-shot text-based retrieval.
Zero-shot retrieval and detection have also been studied in~\cite{dalton2013zero,habibian2014composite,wu2014zero,kim2015ranking}, but no other work has studied zero-shot text-based retrieval in the fine-grained context of CUB and flowers.

There has been a surge of progress in the field of deep multi-modal representation learning in the past several years.
In~\cite{NKKNLN11}, audio and video signals were combined in an autoencoder framework, yielding improved speech signal classification for noisy inputs, and learning a shared representation across modalities.
In~\cite{SS14}, a deep Boltzmann machine architecture was used for multimodal learning on Flickr images and text tags.
In addition to improved discriminative performance, it was also able to hallucinate missing modalities, \ie generate text tags given the image, or retrieve images given text tags.
In~\cite{SSL14}, a novel information theoretic objective is developed, improving the performance of deep multimodal learning for images and text.

Recent image and video captioning models~\cite{mao2014deep,VTBE15,KL15,xu2015show,DHGRVSD15} go beyond tags to generate natural language descriptions.
These models use LSTMs~\cite{HS97} for modeling captions at word level and focus on generating general high-level visual descriptions of a scene.
As an alternative to using LSTMs for language modeling, other works have used character-based convolutional networks~\cite{ZL15}.

Architecturally, other vision systems have trained convolutional and recurrent components (CNN-RNN) end-to-end, e.g. for encoding spatial dependencies in segmentation~\cite{zheng2015conditional} and video classification~\cite{yue2015beyond}.
Here we extend CNN-RNN to learn a visual semantic embedding ``from scratch'' at the character level, yielding competitive performance, robustness to typos, and scalability to large vocabulary. 

A related line of work has been to improve label embeddings for image classification~\cite{BWG10,WBU10,FCSBM13,APHS15,NMBSSFCD13}.
Embedding labels in an euclidean space is an effective way to model latent relationships between classes~\cite{BWG10,WBU10}.
For zero-shot learning, DeViSE~\cite{FCSBM13} and ALE~\cite{APHS15} employ two variants of a ranking formulation to learn a compatibility between images and textual side-information.
ConSe~\cite{NMBSSFCD13}  uses the probabilities of a softmax-output layer to weigh the semantic vectors of all the classes.
Akata et al.~\cite{ARWLS15} showed a large performance gap in zero-shot classification between attributes and unsupervised word embeddings.

In~\cite{elhoseiny2013write} and ~\cite{ba2015predicting}, the zero-shot recognition problem is cast as predicting parameters of a classifier given a text description of the novel category.
Our work considers a similar problem, but there are major differences.
We consider multi-class zero-shot recognition and retrieval, whereas those works mainly focus on one-vs-rest detection of novel categories.
More importantly, our setting assumes that we have a significant amount of visual descriptions for training high-capacity text models, whereas those works had much less text available and used TF-IDF features.

Our contribution builds on previous work on character-level language models~\cite{ZL15} and fine-grained zero-shot learning~\cite{APHS15} to train high capacity text encoders from scratch to jointly embed fine-grained visual descriptions and images.
We demonstrate that with sufficient training data, text-based label embeddings can outperform the previous attributes-based state-of-the art for zero-shot recognition on CUB (at both word and character level).
%
\section{Deep Structured Joint Embedding}
\label{sec:ds_sje}
In this section we describe our approach to jointly embedding images and fine-grained visual descriptions, which we call deep structured joint embedding.
As in previous multimodal structured learning methods~\cite{APHS15,ARWLS15}, we learn a compatibility function of images and text.
However, instead of using a bilinear compatibility function we use the inner product of features generated by deep neural encoders.
An instantiation of our model using a word-level LSTM is illustrated in Figure~\ref{fig:tcnn_model}.
Intuitively, we maximize the compatibility between a description and its matching image, and minimize compatibility with images from other classes.
\vspace{-0.1in}
\paragraph{Objective.} 
Given data $\mathcal{S} = \{(v_{n},t_{n},y_{n}), n = 1, ..., N\}$ containing visual information $v \in \mathcal{V}$, text descriptions $t \in \mathcal{T}$ and class labels $y \in \mathcal{Y}$, we seek to learn functions $f_v : \mathcal{V} \rightarrow \mathcal{Y}$ and $f_t : \mathcal{T} \rightarrow \mathcal{Y}$ that minimize the empirical risk
\begin{align}
\label{eq:objective}
\dfrac{1}{N}\sum_{n=1}^{N}\Delta(y_{n},f_v(v_n)) + \Delta(y_{n},f_t(t_n))
\end{align}
where $\Delta : \mathcal{Y} \times \mathcal{Y} \rightarrow \mathbb{R}$ is the 0-1 loss.
Note that $N$ is the number of image and text \emph{pairs} in the training set, and so a given image can have multiple corresponding captions.
Here we draw a distinction between our method from previous work on structured joint embedding~\cite{ARWLS15}; namely that our objective is symmetric with respect to images and text.
This has the benefit that by optimizing equation~\ref{eq:objective}, a single model can learn to predict by conditioning on both images and text.
We thus name the above objective \emph{deep symmetric structured joint embedding (DS-SJE)}.
It is possible to use just one of the two terms in Eq.~\ref{eq:objective}.
For example in~\cite{ARWLS15} only the first term is used in order to train a zero-shot image classifier, \ie only image encoder $f_{v}$ is trained.
In our experiments we refer to this as deep asymmetric structured joint embedding (DA-SJE).

It is also possible to build an asymmetric model in the opposite direction, \ie only train $f_{t}$ in order to perform zero-shot image retrieval, although we are not aware of previous works doing this.
From a practical perspective it is clearly better to have a single model that does both tasks well.
Thus in our experiments we compare DS-SJE with DA-SJE (training only $f_v$) for zero-shot classification.
\vspace{-0.1in}
\paragraph{Inference.}
We define a compatibility function $F : \mathcal{V} \times \mathcal{T} \rightarrow \mathbb{R}$ that uses features from learnable encoder functions $\theta(v)$ for images and $\varphi(t)$ for text:
\begin{align}
\label{eq:compatibility}
F(v, t) = \theta(v)^T\varphi(t)
\end{align}
We then formulate image and text classifiers as follows:
\begin{align}
\label{eq:classifiers}
f_v(v) = \underset{y \in \mathcal{Y}}{\text{arg max }} \mathbb{E}_{t \sim \mathcal{T}(y)}[F(v, t)]\\
f_t(t) = \underset{y \in \mathcal{Y}}{\text{arg max }} \mathbb{E}_{v \sim \mathcal{V}(y)}[F(v, t)]
\end{align}
where $\mathcal{T}(y)$ is the subset of $\mathcal{T}$ from class $y$, $\mathcal{V}(y)$ is the subset of $\mathcal{V}$ from class $y$, and the expectation is over text descriptions sampled uniformly from these subsets.
Since the compatibility function is shared by $f_t$ and $f_v$, in the symmetric objective it must learn to yield accurate predictions for both classifiers.
From the perspective of the text encoder, this means that text features must produce a higher compatibility score to a matching image compared to both 1) the score of that image with any mismatching text, and 2) the score of that text with any mismatching image.
We found that both 1) and 2) are important for accurate recognition and retrieval using a single model.
\vspace{-0.1in}
\paragraph{Learning.} 
Since the 0-1 loss is discontinuous, we instead optimize a surrogate objective function (related to equation~\ref{eq:objective}) that is continuous and convex:
\begin{align}
\label{eq:objective_actual}
\dfrac{1}{N}\sum_{n=1}^{N} \ell_v(v_n, t_n, y_n) + \ell_t(v_n, t_n, y_n)
\end{align}
where the misclassification losses are written as:
\begin{align}
\ell_v(v_n, t_n, & y_n) = \\
  \underset{y \in \mathcal{Y}}{\max}(0,&\Delta(y_n, y) + \mathbb{E}_{t \sim \mathcal{T}(y)} [ F(v_n,t) - F(v_n,t_n) ])\nonumber\\
\ell_t(v_n, t_n, & y_n) = \\
\underset{y \in \mathcal{Y}}{\max}(0,&\Delta(y_n, y) + \mathbb{E}_{v \sim \mathcal{V}(y)} [ F(v,t_n) - F(v_n,t_n)])\nonumber
\end{align}

In practice we have many visual descriptions and many images per class.
During training, in each mini-batch we first sample an image from each class, and then sample one of its ten corresponding captions.
To train the model, we use SGD on Eq.~\ref{eq:objective_actual} with RMSprop.
Since our text encoder models are all differentiable, we backpropagate (sub)-gradients through all text network parameters for end-to-end training.
For the image encoder, we keep the network weights fixed to the original GoogLeNet.
%
%
%
\section{Text encoder models}
\label{sec:textenc}
In this section we describe the deep neural language models that we use for representing fine-grained visual descriptions.
We compare the performance on zero-shot prediction tasks in Section~\ref{sec:exp}.
\subsection{Text-based ConvNet (CNN)}
Text-based convolutional neural networks were studied in depth in~\cite{ZL15} for the task of document classification.
The text-based CNN can be viewed as a standard CNN for images, except that the image width is 1 pixel and the number of channels is equal to the alphabet size.
The 2D convolution and spatial max-pooling are replaced by temporal (1D) convolution and temporal max-pooling. 
After each convolution layer, we use rectified linear activation unit (ReLU), which is defined as $relu(x) = \max(0,x)$.
The overall network is constructed using convolution, pooling and thresholding activation function layers, followed by fully-connected layers to project onto the embedding space. 
The text embedding function is thus simply $\varphi(t) = \text{CNN}(t)$; the final hidden layer of the CNN.

The maximum input length for character sequences is constrained by the network architecture, but variable length sequences beneath this limit are handled by zero-padding the input past the final input character.
The Word-CNN is exactly the same as Char-CNN except that the alphabet of the Char-CNN is replaced with the vocabulary of the Word-CNN.
Of course, the vocabulary is much larger, typically at least several thousand words compared to a few dozen characters in an alphabet.
However, the sequence length is significantly reduced.
\subsection{Convolutional Recurrent Net (CNN-RNN)}
\label{sec-cnnrnn}
A potential shortcoming of convolution-only text models is that they lack a strong temporal dependency along the input text sequence.
However, the CNN models are extremely fast and scale well to long sequences such as character strings.
To get the benefits of both recurrent models and CNNs, we propose to stack a recurrent network on top of a mid-level temporal CNN hidden layer.
Intuitively, the CNN hidden activation is split along the time dimension (in our case when the dimension was reduced to 8) and treated as an input sequence of vectors.
The entire resulting network is still end-to-end differentiable.

This approach has the advantage that low-level temporal features can be learned efficiently with fast convolutional networks, and temporal structure can still be exploited at the more abstract level of mid-level features.
This can be viewed as modeling temporal structure at the abstract or conceptual level, not strictly dilineated by word boundaries.
%
The approach is well-suited to the case of character-level processing (Char-CNN-RNN). We also evaluate a word-level version (Word-CNN-RNN).

\begin{figure}[t!]
\begin{center}
\includegraphics[width=0.98\columnwidth]{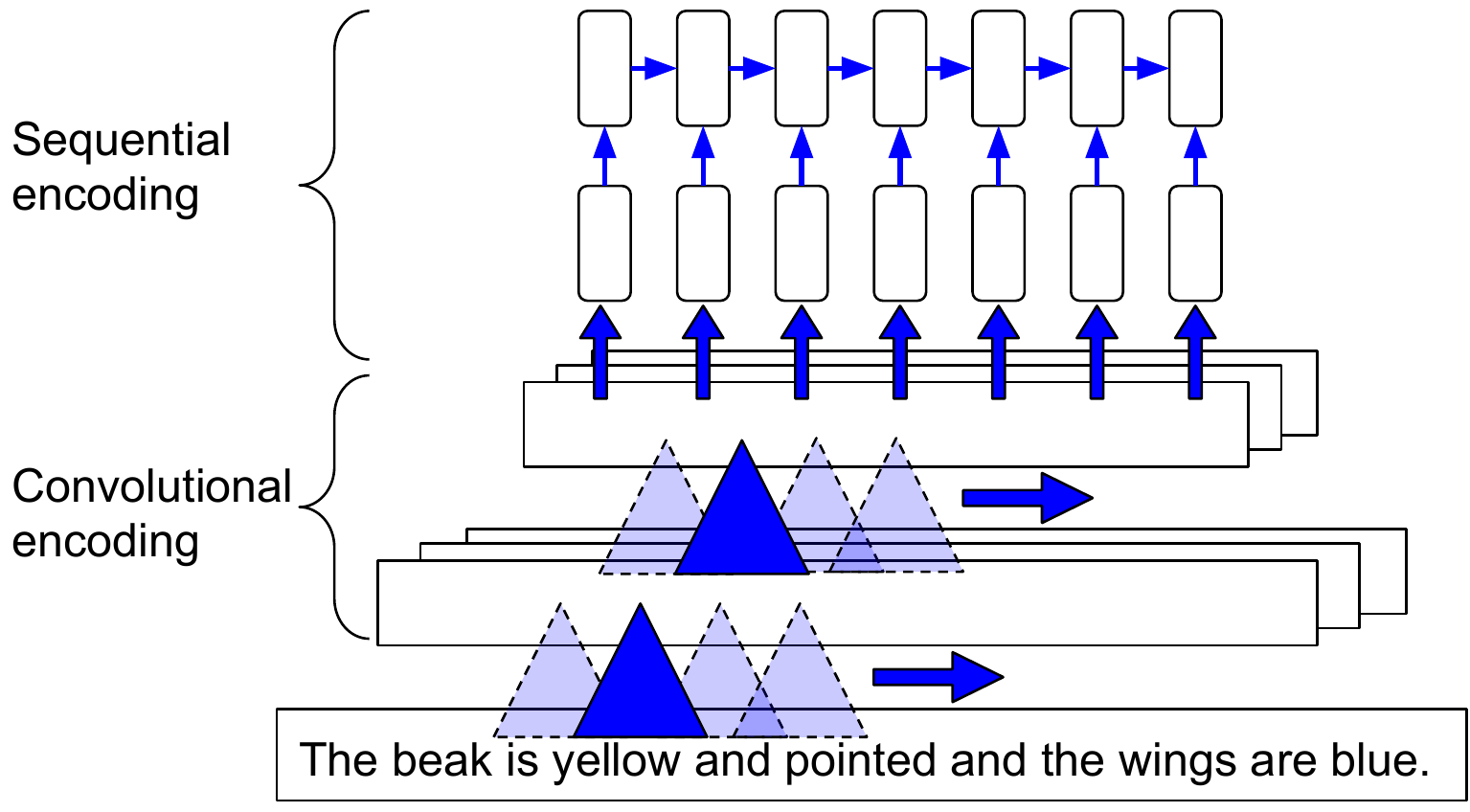}
\end{center}
\vspace*{-0.15in}
\caption{Our proposed convolutional-recurrent net.}
\vspace*{-0.1in}
\label{fig:hybrid}
\end{figure}

Figure~\ref{fig:hybrid} illustrates the convolutional-recurrent approach.
%
The final encoded feature is the average hidden unit activation over the sequence, \ie $\varphi(t) = 1/L \sum_{i=1}^{L} h_{i}$, where $h_i$ is the hidden activation vector for the $i$-th frame and $L$ is the sequence length.
The resulting scoring function can be viewed as a linear accumulation of evidence for compatibility with a query image (illustrated in Figure~\ref{fig:tcnn_model}).
It is also a linearized version of attention over the text sequence.
This has the advantage that at test time for classification or retrieval, one can use the averaged hidden units as a feature, but for diagnostic purposes one can backtrace the score computation to each time step of text processing.
\subsection{Long Short-Term Memory (LSTM)}
As opposed to the CNN models, the LSTM explicitly takes into account the temporal structure starting from words or characters.
We refer readers to~\cite{HS97} for full details.
To extract a text embedding from the LSTM text encoder, we take the temporal average of the final layer hidden units, i.e. $\varphi(t) = 1/L \sum_{i=1}^{L} h_{i}$ (defined similarly as in Section~\ref{sec-cnnrnn}).
%
%
\subsection{Baseline representations}
Since we gathered a significant amount of new data, traditional (\eg non-``deep'') text representations should also improve in performance.
To evaluate whether using the neural language models really provide an additional benefit, we compare against several classical methods.
For the BoW model, we first compute the vocabulary $V$ of all of the unique words appearing in the visual descriptions.
Then, we encode each description as a binary vector indicating the presence or absence of each word.
The embedding function is simply the output of a multi-layer perceptron (MLP), $\varphi(t) = \text{MLP}(I(t))$.
where $I(\cdot)$ maps $t$ to an indicator vector in $\{0,1\}^{|V|}$.
In practice we found a single layer linear projection was sufficient for surprisingly good performance.

We also evaluate a baseline that represents descriptions using unsupervised word embeddings learned by word2vec~\cite{MSCCD13}.
Previous works on visual-semantic embedding have directly used the word embedings of target classes for zero-shot learning tasks.
However, in our case we have access to many visual descriptions, and we would like to extract vector representations of them in real time; \ie without re-running word2vec training.
A very simple way to do this is to average the word embeddings of each word in the visual description.
Although this loses the structure of the sentence, this nevertheless yields a strong baseline and in practice performs similarly to bag of words.

Finally, an important point of comparison is attributes, which contain rich structured information far more compactly than informal visual descriptions.
As in the case of bag-of-words, we learn a single-layer encoder function mapping attributes to the embedding space.
Since the number of attribute vectors is very small (only one per class), the risk of over-fitting strongly limits the encoder network capacity.
The CUB dataset also has per-image attributes, but we found that using these does not improve performance compared to using a single averaged attribute vector per class. 
\section{Experimental results}
\label{sec:exp}
\begin{figure}[t!]
\begin{center}
\includegraphics[width=\linewidth, trim=0 0 320 0,clip]{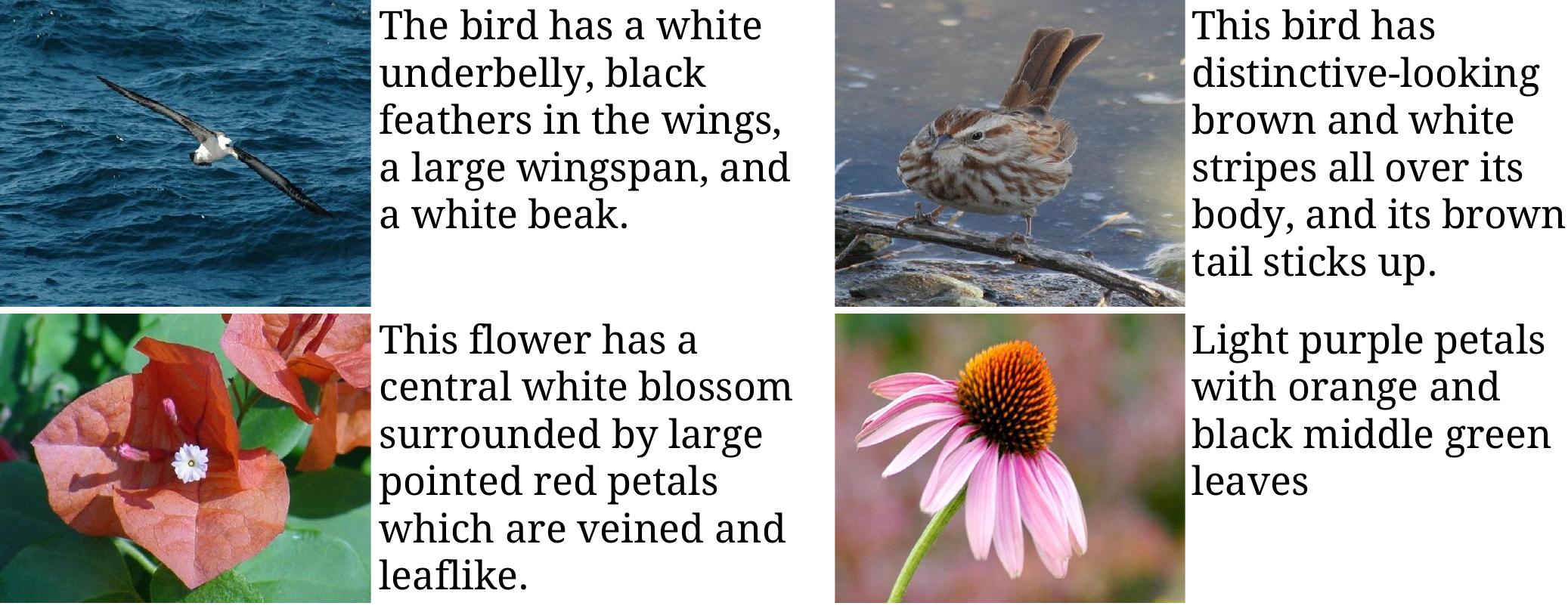}
\end{center}
\vspace{-0.25in}
\caption{Example annotations of birds and flowers.}
\label{fig:example_bird_data}
\vspace{-0.2in}
\end{figure}
In this section we describe our experiments on the Caltech-UCSD Birds dataset (CUB) and Oxford Flowers-102 (Flowers) dataset.
CUB contains 11,788 bird images from 200 different categories. Flowers contains 8189 flower images from 102 different categories.
Following~\cite{APHS15}, the images in CUB are split into 100 training, 50 validation, and 50 disjoint test categories\footnote{Since we evaluate in the zero-shot setting, it is critical that the validation categories be disjoint from the training categories. Once hyperparameters have been cross-validated, the training + validation (150) classes can be taken as the training set. For Flowers, we do not do any parameter cross-validation, we use the same parameters found for CUB.}.
As in~\cite{ba2015predicting}, the images in Flowers are split into 82 training + validation and 20 test classes.
For the image features, we extracted $1,024$-dimensional pooling units from GoogLeNet~\cite{szegedy2015going} with batch normalization~\cite{ioffe2015batch} implemented in Torch\footnote{\url{github.com/soumith/imagenet-multiGPU.torch}}.
For each image, we extracted middle, upper left, upper right, lower left and lower right crops for the original and horizontally-flipped image, resulting in 10 views per training image.
At test time we only use the original image resized to $224 \times 224$.

For all word-level models (BoW, Word-LSTM, Word-CNN, Word-CNN-RNN), we used all vocabulary words in the dataset. 
For character-level models (Char-LSTM, Char-CNN, Char-CNN-RNN), the alphabet consisted of all lowercase characters and punctuation.

The CNN input size (sequence length) was set to $30$ for word-level and $201$ for character-level models; longer text inputs are cut off at this point and shorter ones are zero-padded.
All text embeddings used a 1024-dimensional embedding layer to match the size of the image embedding.
We kept the image encoder fixed, and used RMSprop with base learning rate $0.0007$ and minibatch size $40$. 

\subsection{Collecting fine-grained visual descriptions}
\label{sec:fg-vd}
In this section we describe the collection of our new dataset of fine-grained visual descriptions.
For each image in CUB and Flowers, we collected ten single-sentence visual descriptions.
We used the Amazon Mechanical Turk (AMT) platform for data collection, using non-``Master'' certified workers situated in the US with average work approval rating above $95\%$.

We asked workers to describe only visual appearance in at least 10 words, to avoid figures of speech, to avoid naming the species even if they knew it, and not to describe the background or any actions being taken.
The prompt included three example sentences and a diagram labeling specific parts of a bird (e.g. tarsus) and flower (e.g. stamen) so that non-experts could describe many different aspects without reference to external sources such as Wikipedia.
Workers were not told the species.
Figure~\ref{fig:example_bird_data} shows several representative examples of the results from our data collection.
The descriptions almost always accurately describe the image, to varying degrees of comprehensiveness.
Thus, in some cases multiple captions might be needed to fully disambiguate the species of bird category.
However, as we show subsequently, the data is descriptive and large enough to support training high-capacity text models and greatly improve the performance of text-based embeddings for zero-shot learning.

\begin{table}[t]
  \centering
  \resizebox{\linewidth}{!}{
    \begin{tabular}{l|c|c|c|c|}
    \cline{2-5}
    & \multicolumn{2}{|c|}{Top-1 Acc (\%)}
       & \multicolumn{2}{|c|}{AP@50 (\%)} \\
    \hline
    \multicolumn{1}{|l|}{\textbf{Embedding}} & \textbf{DA-SJE} & \textbf{DS-SJE} & \textbf{DA-SJE} & \textbf{DS-SJE} \\
    \hline
    \multicolumn{1}{|l|}{\textsc{Attributes}}  & $50.9$ & $50.4$  & $20.4$ & $\mathbf{50.0}$\\
    \hline
    \multicolumn{1}{|l|}{\textsc{Word2Vec}} & $38.7$ & $38.6$  & $7.5$ & $33.5$ \\
    \hline
    \multicolumn{1}{|l|}{\textsc{Bag-of-Words}}  & $43.4$ & $44.1$ & $24.6$ & $39.6$  \\
    \hline \hline
    \multicolumn{1}{|l|}{\textsc{Char Cnn}}  & $47.2$ & $48.2$  & $2.9$ & $42.7$\\
    \hline
    \multicolumn{1}{|l|}{\textsc{Char Lstm}}  & $22.6$ & $21.6$  & $11.6$ & $22.3$\\
    \hline
    \multicolumn{1}{|l|}{\textsc{Char Cnn-Rnn}}  & $54.0$ & $54.0$ & $6.9$ & $45.6$ \\
    \hline \hline
    \multicolumn{1}{|l|}{\textsc{Word Cnn}}  & $50.5$ & $51.0$  & $3.4$ & $43.3$\\
    \hline
    \multicolumn{1}{|l|}{\textsc{Word Lstm}}  & $52.2$ & $53.0$  & $\mathbf{36.8}$ & $46.8$\\
    \hline
    \multicolumn{1}{|l|}{\textsc{Word Cnn-Rnn}}  & $\mathbf{54.3}$ & $\mathbf{56.8}$  & $4.8$ & $48.7$\\
    \hline
    \end{tabular}
  }
  \caption{Zero-shot recognition and retrieval on CUB. ``DS-SJE'' and ``DA-SJE'' refer to symmetric and asymmetric forms of our joint embedding objective, respectively.}
  \label{tab:fivesentences_CUB}
 \end{table} 

\begin{figure*}[ht!]
\centering
\begin{subfigure}{0.46\linewidth}
   \includegraphics[width=0.49\linewidth, trim=40 0 45 70]{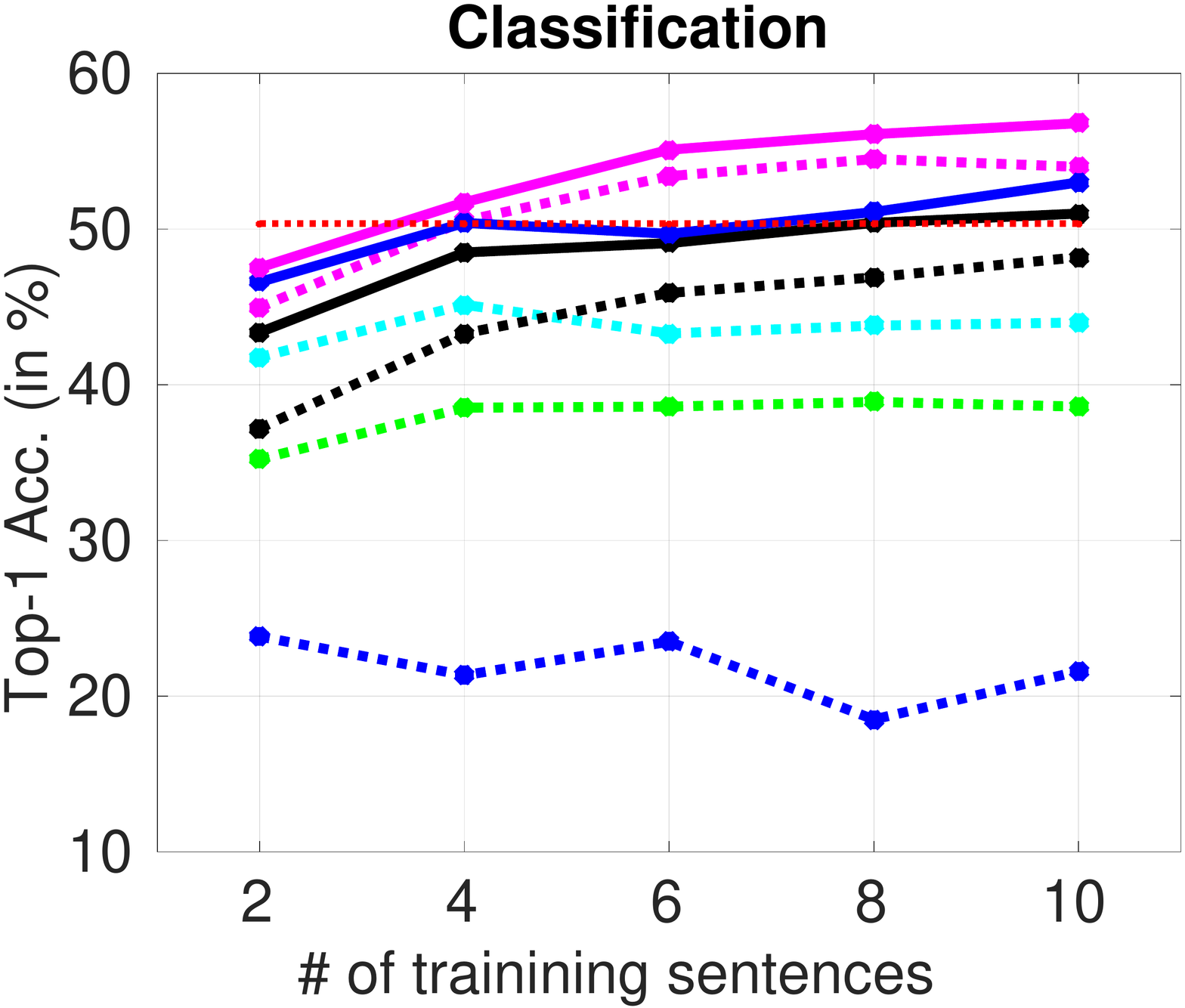}
   \includegraphics[width=0.49\linewidth, trim=40 0 45 70]{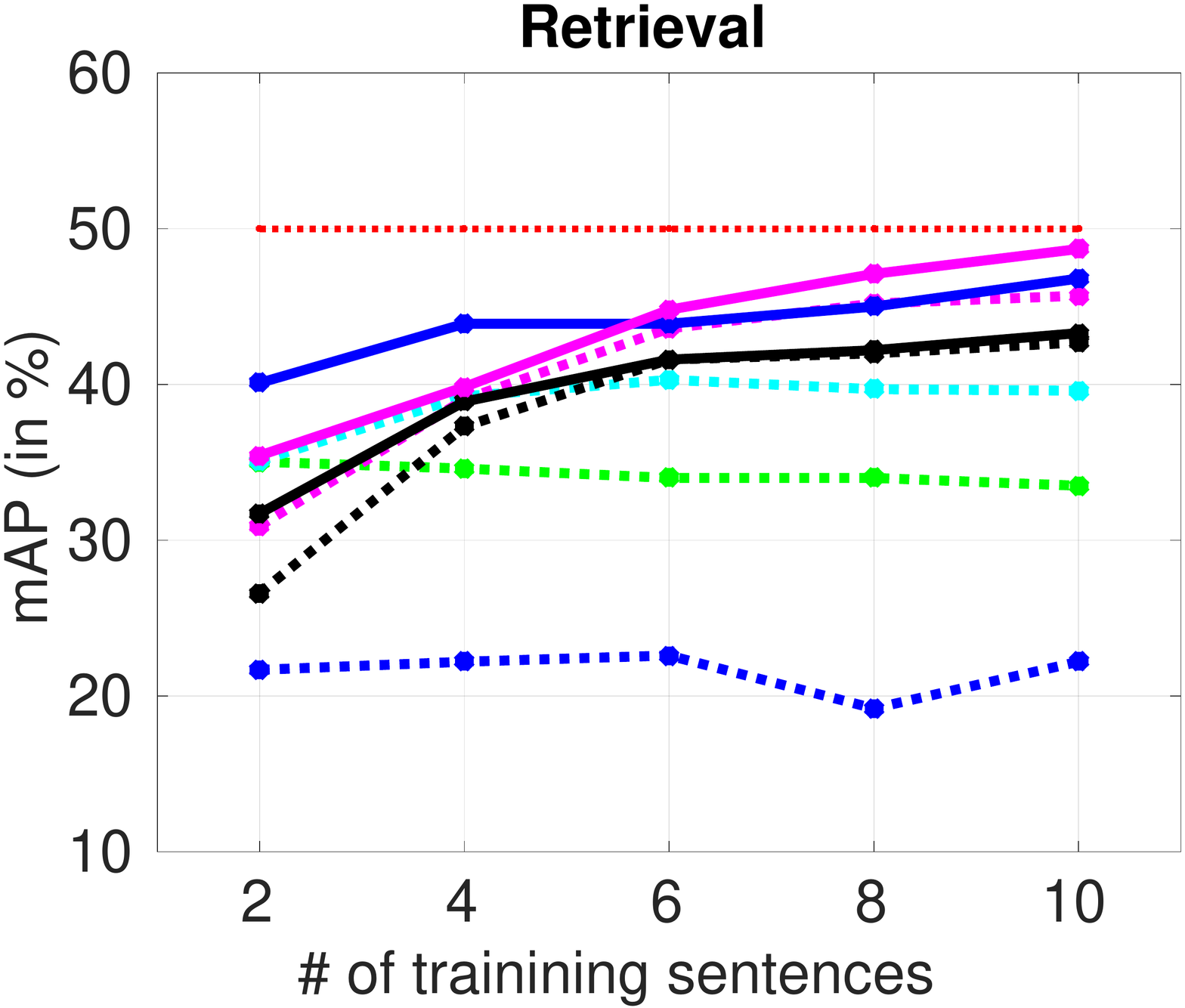}
\caption{Increasing number of training sentences}
\label{fig:numtrsentence}
\end{subfigure}
\hfill
\begin{subfigure}{0.46\linewidth}
   \includegraphics[width=0.49\linewidth, trim=40 0 45 70]{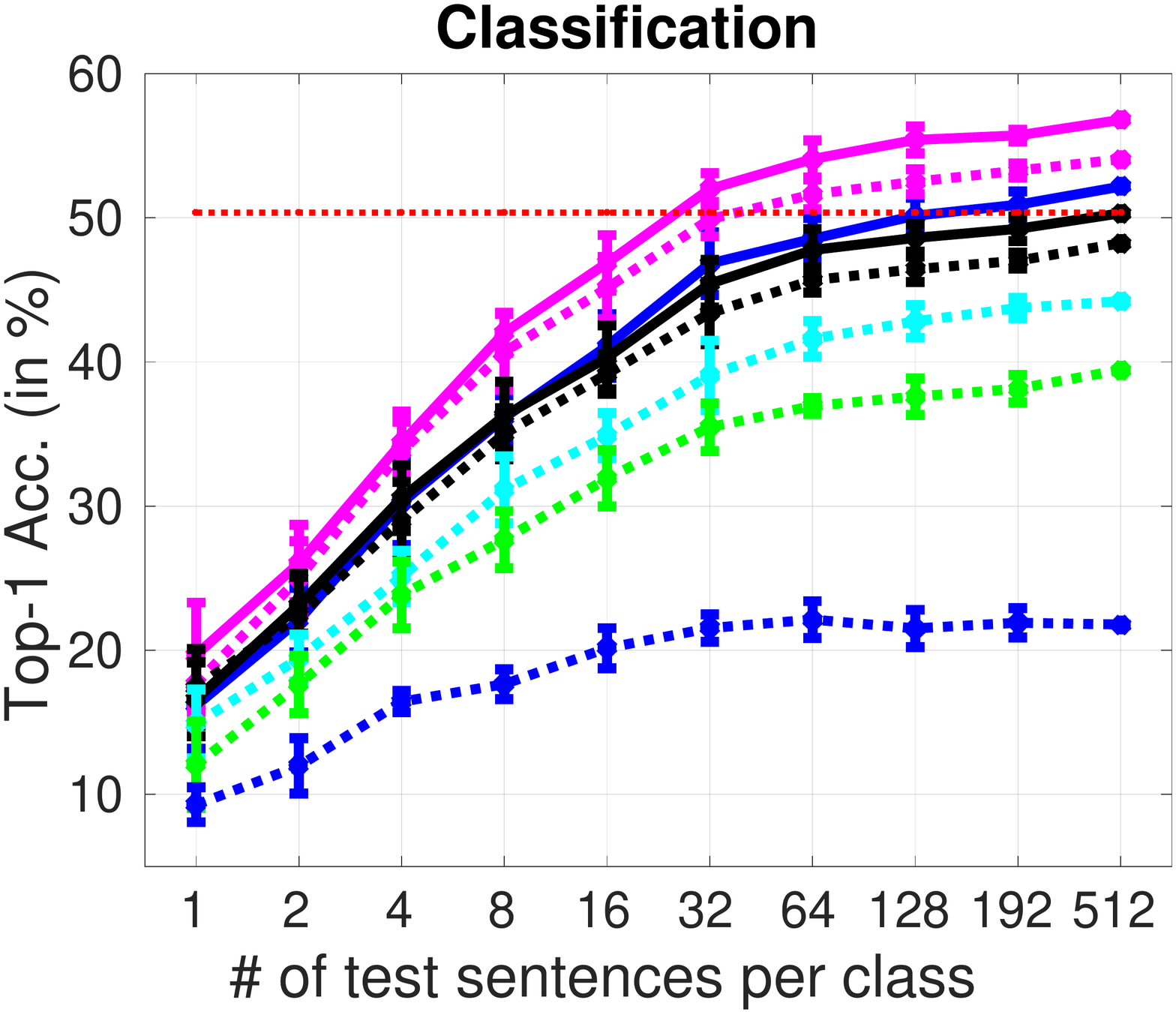}
   \includegraphics[width=0.49\linewidth, trim=40 0 45 70]{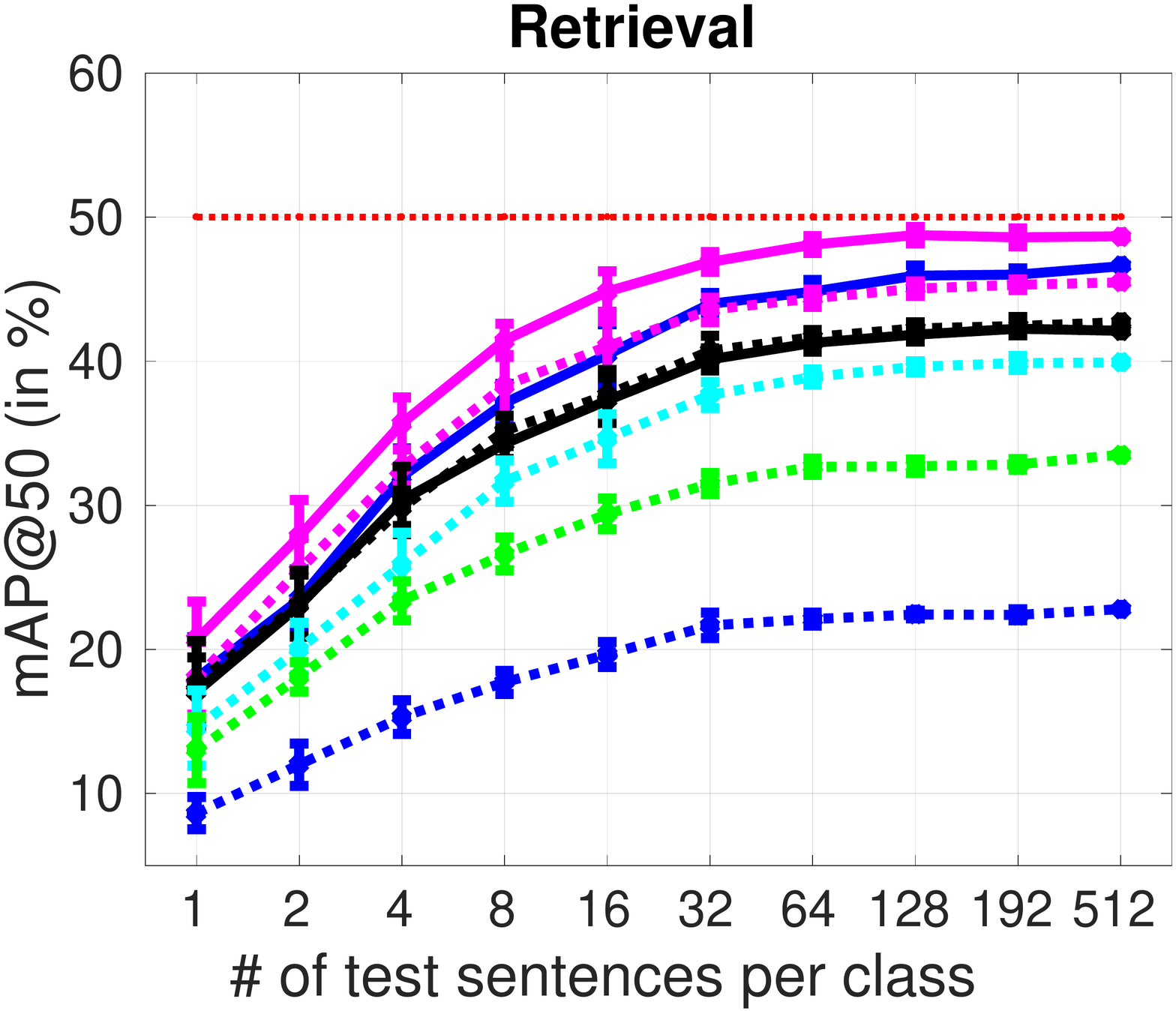}
\caption{Increasing number of test sentences}
\label{fig:numtstsentence}
\end{subfigure}
\hfill
\begin{subfigure}{0.07\linewidth}
   \includegraphics[width=\linewidth, trim=40 0 30 40]{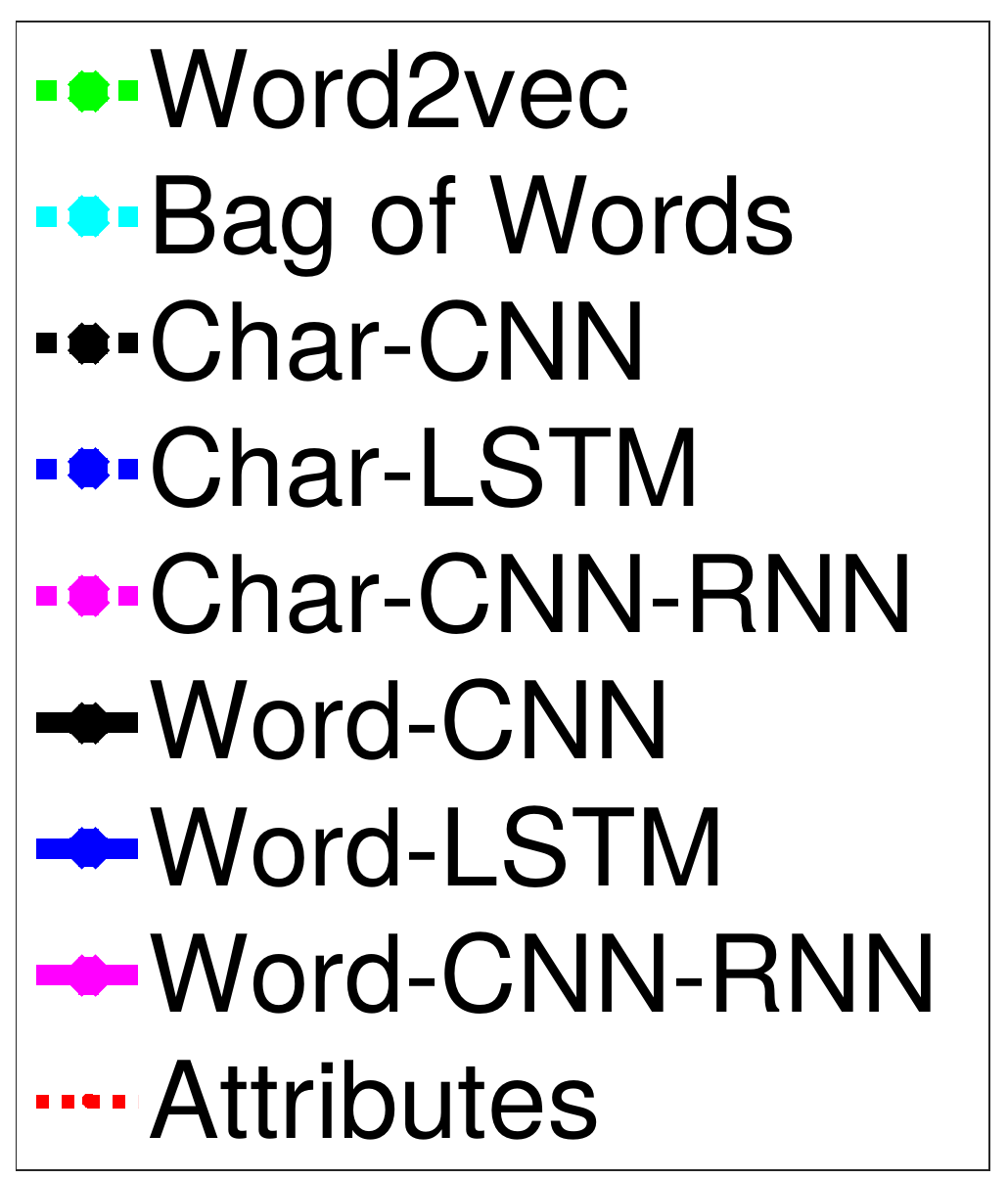}
\end{subfigure}
\vspace*{-0.1in}
\caption{Zero-shot image classification and retrieval accuracy versus number of sentences per-image used in training and number of sentences in total used for testing. Results reported on CUB.  }
\vspace*{-0.1in}
\label{fig:numsentence}
\end{figure*}

\subsection{CUB zero-shot recognition and retrieval}
\label{sec:CUB}
In this section we describe the protocol and results for our zero-shot tasks.
For both recognition and retrieval, we first extract text encodings from test captions and average them per-class.
In this experiment we use \emph{all} test captions and in a later section we vary this number, including using a single caption per class.
In recognition, the resulting classifier is defined by equation~\ref{eq:classifiers}.
Note that by linearity we can move the expectation inside the compatibility function:
\begin{align}
f_v(v) = \underset{y \in \mathcal{Y}}{\text{arg max }} \theta(v)^{T} \mathbb{E}_{t \sim \mathcal{T}(y)}[\varphi(t)]
\end{align}
The expectation above is estimated by the averaged per-class text embedding that we compute.
Hence the accuracy of the classifier is determined not only by the underlying image and text encoders, but also by the quantity of text available at test time to improve the estimate.

In the retrieval task, we rank all test set images according to compatibility (equation~\ref{eq:compatibility}) with the averaged text embedding for each class.
We report the AP@50, \ie the percent of top-50 scoring images whose class matches that of the text query, averaged over the 50 test classes.

Table~\ref{tab:fivesentences_CUB} summarizes our results. 
Both in the classification (first two columns) and for retrieval (last two columns) settings,  the 
symmetric (DS-SJE) formulation of our model improves over the asymmetric (DA-SJE) formulation. 
Especially for retrieval, DS-SJE performs much better than DA-SJE consistently for all the text embedding variants.
It makes the difference between working very well and failing, particularly for the high-capacity models which likely overfit to the classification task in the asymmetric setting.

In the classification setting there are notable differences between the language models. 
For DA-SJE (first column), Char-CNN-RNN (54.0\% Top-1 Acc) and Word-CNN-RNN (54.3\%) outperform the attributes-based state-of-the-art~\cite{ARWLS15} for zero-shot classification (50.1\%).
In fact we replicated the attribute-based model in~\cite{ARWLS15} and got slightly better results (50.9\%, also reported in Table~\ref{tab:fivesentences_CUB}), probably due to training on 10 image crops instead of a single crop.
Similar observations hold for DS-SJE (second column). 
Notably for DS-SJE, Char-CNN-RNN (54.0$\%$), Word-CNN (51.0$\%$), Word-LSTM (53.0$\%$) and Word-CNN-RNN (56.8$\%$) outperform the attributes. 
In the case of retrieval and DS-SJE (last column), attributes still performs the best (50.0\% AP), but Word-CNN-RNN (48.7\%) approaches this result.

Among the character-level models, Char-CNN is significantly better than Char-LSTM.
Additionally, our proposed Char-CNN-RNN, which adds a temporal aspect to Char-CNN, improves over the other two character-based deep methods and also over the attribute-based state-of-the-art for classification.
This is notable because it establishes that character-level models can extract visually-discriminative text representations of previously-unseen categories.
Furthermore, combining convolutional and temporal processing appears to be a promising approach to learn at the character level.
Word-level models improve performance further and can also significantly outperform attributes. 

\subsection{Effect of visual description training set size}
\label{sec:increasingsentences}
In this section we investigate the effect of increasing the number of sentences used in training on zero-shot classification and retrieval performance. 
Obviously having more data is better, but with this experiment we can see which methods are best at which operating point of data size (hence cost).
We start with using one sentence per image and we increase this number gradually to ten sentences per image for training.
For testing, the protocol is the same as in Table~\ref{tab:fivesentences_CUB}, and we use all available captions per class.
We show the performance of several text encoding models in Fig~\ref{fig:numtrsentence}. 
In zero-shot classification, attributes are competitive when two captions per-image are available, but with more training captions the deep network models win.
For retrieval, the crossover point might happen with more than ten captions per image as the results seem to be increasing.
The baseline word2vec and BoW encodings do not gain much from more data.
The results suggests that given a moderate number of sentences, \ie four per image, neural text encoders improve the performance over the state-of-the-art attribute-based methods significantly.
Among neural text encoders, Char-LSTM fares worst and also does not appear to gain consistently from additional data.
It may be that the long training sequence length increases the difficulty of LSTM training, relative to the word-based approach.
Stacking a recurrent module on top of a text convolutional network appears to avoid this problem, achieving significantly better performance than the Word-LSTM especially with more than 4 sentences for training.
It also has the nice property of robustness to typos.
Overall, Word-CNN-RNN achieved the best performance.
\subsection{Effect of test visual description length}
\label{sec:variable_text}
In a real application relating images and text (\eg text-based image retrieval), most users would prefer to describe a visual concept concisely, rather than writing a detailed article with many sentences.
Thus, we evaluate the performance of our model using a varying number of query descriptions per class at test time.
The experimental protocol is a slight modification of that used in Table~\ref{tab:fivesentences_CUB}.

As before, we extract text embeddings from test set captions and average them per-class.
In this case, we extract embeddings separately using $\{1,2,4,8,16,32,64,128\}$ and also \emph{all} descriptions available per class.
For each description length, we report the resulting zero-shot classification accuracy and zero-shot retrieval AP$@50$.
Since we do not use all available test captions per class, we perform $10$ iterations of this procedure while randomly sampling the descriptions used for each class.

Figure~\ref{fig:numtstsentence} shows the averaged results for zero-shot classification and for zero-shot retrieval.
Both figures include error bars to $\pm 1$ standard deviation.
%
Note that the error bars are larger towards the left side of both figures because in the few-text case, especially discriminative or especially vague (or wrong) descriptions can have a relatively larger impact on the text embedding quality.
BoW again shows a surprisingly good performance, significantly better than word2vec and competitive with Char-CNN.
However, the word-level neural text encoders outperform word2vec and BoW at all operating points.
%

%
\begin{table}[t]
  \centering 
  \resizebox{\linewidth}{!}{
    \begin{tabular}{l|c|c|c|c|}
    \cline{2-5}
    & \multicolumn{2}{|c|}{Top-1 Acc (\%)}
       & \multicolumn{2}{|c|}{AP@50 (\%)} \\
    \hline
    \multicolumn{1}{|l|}{\textbf{Embedding}} & \textbf{DA-SJE} & \textbf{DS-SJE} & \textbf{DA-SJE} & \textbf{DS-SJE} \\
    \hline
    \multicolumn{1}{|l|}{\textsc{Word2Vec}} & $54.6$ & $54.2$ & $16.3$ & $52.1$ \\
    \hline
    \multicolumn{1}{|l|}{\textsc{Bag-of-Words}} & $56.7$ & $57.7$ & $28.2$ & $57.3$ \\
    \hline \hline
    \multicolumn{1}{|l|}{\textsc{Char CNN}}  & $51.1$ & $47.3$ & $8.3$ & $46.1$\\
    \hline 
    \multicolumn{1}{|l|}{\textsc{Char Lstm}}  &$29.1$ & $25.8$ & $19.3$ & $27.0$\\
    \hline
    \multicolumn{1}{|l|}{\textsc{Char Cnn-Rnn}}  &$61.7$ & $63.7$ & $13.6$ & $57.3$\\
    \hline \hline
    \multicolumn{1}{|l|}{\textsc{Word Cnn}}  & $60.2$ & $60.7$ & $8.7$ & $56.3$\\
    \hline
    \multicolumn{1}{|l|}{\textsc{Word Lstm}}  & $\mathbf{62.3}$ & $64.5$  & $\mathbf{45.9}$ & $52.3$\\    
    \hline
    \multicolumn{1}{|l|}{\textsc{Word Cnn-Rnn}}  &$60.9$ & $\mathbf{65.6}$  & $7.6$ & $\mathbf{59.6}$\\
    \hline
    \end{tabular}
  }
  \caption{Zero-shot \% recognition accuracy and retrieval average precision on Flowers.}
  \label{tab:fivesentences_Oxf}
\end{table}

\begin{figure*}[t]
\begin{center}
   \includegraphics[width=0.9\linewidth]{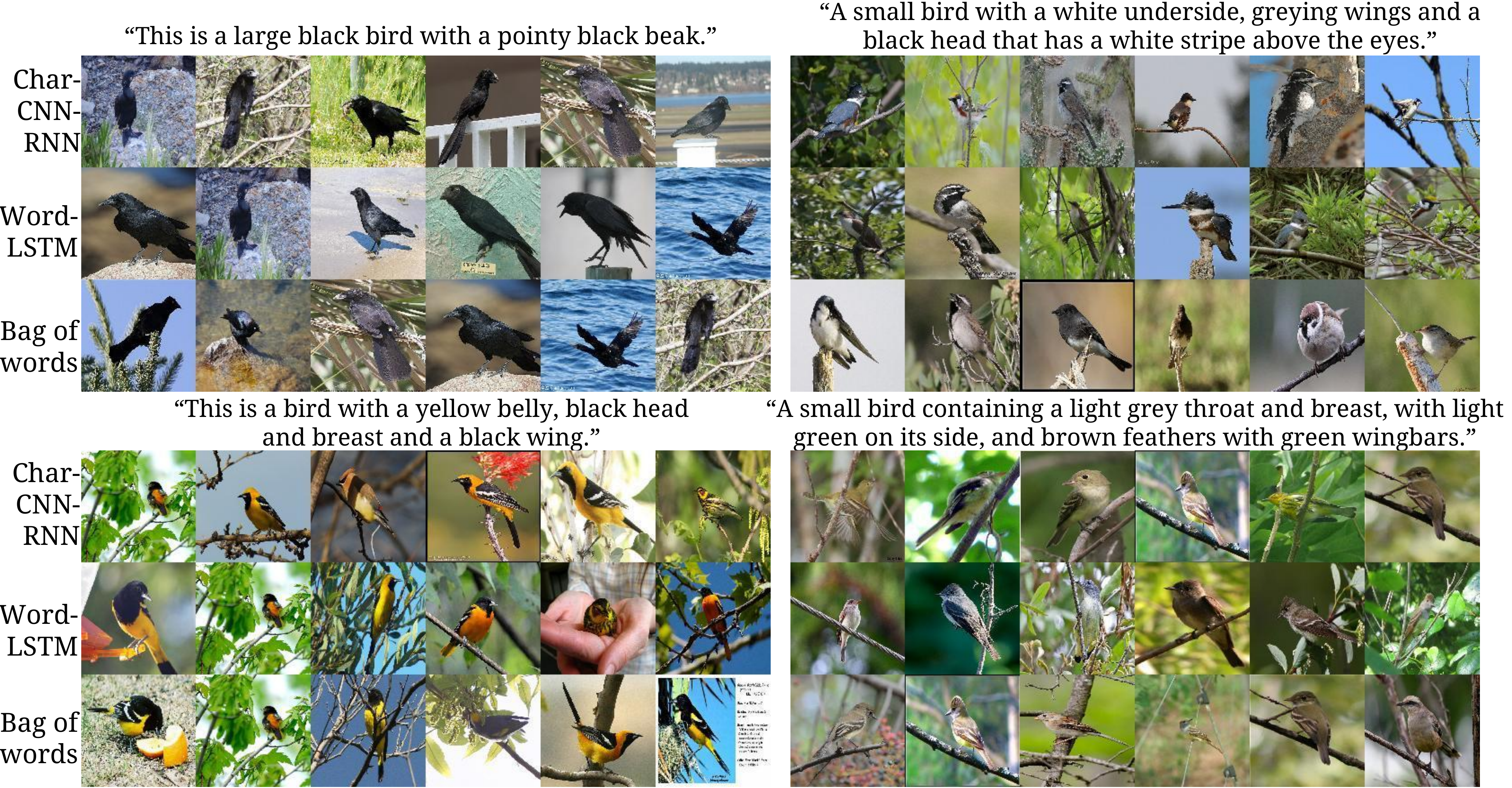}
   \vspace{-0.1in}
   \caption{Zero-shot retrieval given a single query sentence. Each row corresponds to a different text encoder.} 
   \label{fig:retrieval_qualitative}
\end{center}
  \vspace{-0.25in}
\end{figure*}
\subsection{Flowers zero-shot recognition and retrieval}
\label{sec:Oxf}
To demonstrate that our results generalize beyond the case of bird images, we report the same set of experiments on the Flowers dataset.
The experimental setting here is the same as in Sec~\ref{sec:CUB}, except that
there is no attributes baseline due to lack of labeled attributes for this dataset.
All neural text model architectures are the same as we used for CUB, and we used the same hyperparameters from cross-validation on CUB.
Table~\ref{tab:fivesentences_Oxf} summarizes our results.

Char CNN-RNN achieves competitive results to word-level models both for DA-SJE and DS-SJE.
The word-level models achieve the best result, significantly better than both the shallow embeddings and character-level models.
Among different models, Word LSTM is the winner for DA-SJE both in classification and retrieval.
On the other hand, Word CNN-RNN is the winner for DS-SJE for the same.
As in the case for CUB, we found that DS-SJE achieves strong retrieval performance, and DA-SJE often fails in comparison.

\begin{table}[t]
	\centering 
	\begin{small}
		\begin{tabular}{|l|c|c|}
			\hline
			\textbf{Approach} & \textbf{CUB} & \textbf{Flowers} \\    
			\hline
			CSHAP${_H}$~\cite{HEEY15} & $17.5$ & -- \\
			\hline
			AHLE~\cite{APHS15} & $27.3$ & -- \\    
			\hline
			TMV-HLP~\cite{FHXG15} & $47.9$ & -- \\
			\hline        
			SJE~\cite{ARWLS15} & $50.1$ & -- \\
			\hline
			DA-SJE (ours) & $54.3$ & $62.3$ \\
			\hline
			DS-SJE (ours) & $\mathbf{56.8}$ & $\mathbf{65.6}$ \\
			\hline
		\end{tabular}
	\end{small}
	\caption{Summary of zero-shot \% classification accuracies. Note that different features are used in each work, although~\cite{APHS15} uses the same features as in this work.} 
	\label{tab:soa}
\end{table}

\subsection{Qualitative results}
\label{sec:qualitative}

Figure~\ref{fig:retrieval_qualitative} shows several example zero-shot retrieval results using a single text description. 
Both the text queries and images are real data points drawn from the test set.
We observe that having trained on our dataset of visual descriptions, our proposed method returns results that accurately reflect the text, even when using only a single caption.
Quantitatively, BoW achieves $14.6\%$ AP@50 with a single query compared to $18.0\%$ with word-LSTM and $20.7\%$ with Word-CNN-RNN.

\begin{figure}[t]
	\begin{center}
		\includegraphics[width=\columnwidth]{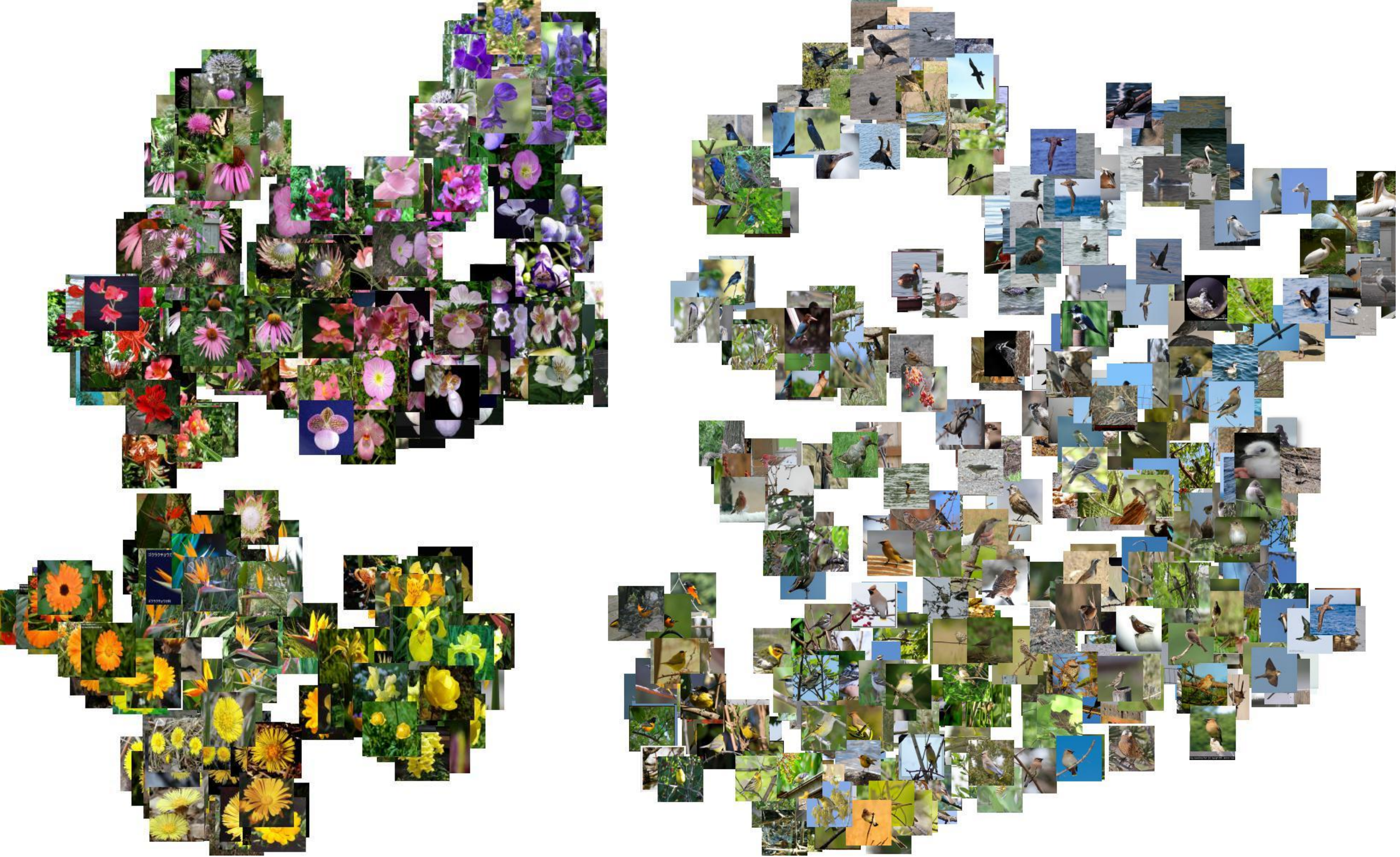}
		\caption{t-SNE embedding of \emph{test class} description embeddings from Oxford-102 (left) and CUB (right), marked with corresponding images. Best viewed with zoom. \label{fig:tsne}}
	\end{center}
	\vspace{-0.3in}
\end{figure}

Note that although almost all retrieved images match the text query well, the actual class of that image can still be incorrect.
This is why the average precision may seem low compared to the generally good qualitative results.
The performance appears to degrade gracefully; our model at least returns visually-consistent results if not of the correct class.
Furthermore, some queries are inherently ambiguous and could match multiple classes equally well, so low precision is not necessarily the fault of the model.
We show a t-SNE embedding of test-set description embeddings in Figure~\ref{fig:tsne}, successfully clustering according to visual similarities (i.e. color, shape).
Additional examples from test images and queries are included in the supplementary material.

\subsection{Comparison to the state-of-the-art}
In this section we compare to the previously published results on CUB, including results that use the same zero-shot split.
CSHAP${_H}$~\cite{HEEY15} uses 4K-dim features from the Oxford VGG net~\cite{Simonyan15} and also attributes to learn a hypergraph on the attribute space.
AHLE~\cite{APHS15} uses Fisher vector image features and attribute embeddings to learn a bilinear compatibility function between these embeddings.
TMV-HLP~\cite{FHXG15} builds a hypergraph on a multiview embedding space learned via CCA which uses deep image features and attributes.
In SJE~\cite{ARWLS15} as in AHLE~\cite{APHS15} a compatibility function is learned, in this case between 1K-dim GoogleNet~\cite{szegedy2015going} features and various other embeddings including attributes.
Our method achieves significant improvements over all of these baselines, despite the fact that we do not use attributes.

Previously-reported zero-shot results on the Flowers dataset~\cite{elhoseiny2013write, ba2015predicting} do not report multi-class classification (instead reporting binary one-vs-rest detection of unseen categories) or do not currently have published splits.
However, it will be interesting to compare these methods of ``predicting a classifier'' given image descriptions in the large-data setting with our new caption collection.
We include our Flowers multi-class results and will publish our split.

Overall, the results in Table~\ref{tab:soa} demonstrate that state-of-the-art zero-shot prediction performace can be achieved directly from text descriptions.
This does not require access to any form of test label embeddings.
Although attributes are richer and more compact than text descriptions, attributes alone form a very small training set.
One explanation for the better performance of using our descriptions is that having many noisy human-generated descriptions acts as an effective regularizer on the learned compatibility function.
This is especially important when training deep networks, which in our model are used for both the image and text encoding components.
Indeed, we observed that when training with attributes, we had to use far fewer epochs (7 compared to 300) to avoid over-fitting.
\vspace{-0.05in}
\section{Discussion}
\vspace{-0.05in}
\label{sec:conc}
We developed a deep symmetric joint embedding model, collected a high-quality dataset of fine-grained visual descriptions, and evaluated several deep neural text encoders. 
We showed that a text encoder trained from scratch on characters or words can achieve state-of-the-art zero-shot recognition accuracy on CUB, outperforming attributes. 
Our text encoders achieve a competitive retrieval result compared to attributes, and unlike attributes can be directly used to build a language-based retrieval system.

Our visual descriptions data also improved the zero shot accuracy using BoW and word2vec encoders.
While these win in the smaller data regime, higher capacity encoders dominate when enough data is available.
Thus our contributions (data, objective and text encoders) improve performance at multiple operating points of training text size.
\vspace{-0.05in}
\subsection*{Acknowledgments}
\vspace{-0.05in}
This work was supported in part by NSF CAREER IIS-1453651, ONR N00014-13-1-0762 and NSF CMMI-1266184.
\vspace{-0.1in}
{\small
\bibliographystyle{ieee}
\bibliography{references}
}
\end{document}